# Evolution of US Oral Political Language

**Jacques Savoy**

Computer Science Dept., University of Neuchatel
rue Emile Argand 11, 2000 Neuchatel (Switzerland)

Jacques.Savoy@unine.ch

**Abstract**

The analysis of US political language is usually based on the written form (e.g. presidential addresses) or posts broadcasted on various social networks. Oral production, however, which is even more frequent, can better reveal the style and mode of thinking of the speaker. This study covers this mode of linguistic communication by considering 19 candidates from the presidential elections between 1960 to 2024. Our main research objectives are to disclose the main trends hidden in those presidential debates. Do we observe a clear simplification of the US political language over time? Does Trump have poor language compared to the other candidates? Do unusual stylistic features occur only with a single, specific president? Moreover, can we detect a pattern explaining the success or failure of some nominees? Over time, this study demonstrates a significant reduction in political language complexity, a decrease of the mean sentence length, and a noteworthy decline of complex terms. Moreover, the emotional tone increases over the decades, while logical and rational thinking tends to lessen.

**Keywords**: political speeches, language complexity evolution, oral communication analysis, stylometry.

## 1 Introduction

Political language is evolving to better convince citizens that the proposed political offer is the most pertinent one for the nation. Previous studies to analyse such a development are grounded on the written form that is easily accessible and relatively frequent. This communication mode corresponds to official speeches, which are first written and then pronounced in front of an audience such as that of Parliament (Gennaro et al. 2021). This choice, however, usually reflects collaborative work between the speaker (e.g. chief executive or policymaker) and staffers (Rule et al. 2015; Savoy 2025). Thus, such a written mode does not precisely echo the personal style and rhetoric of a politician.

Based on transcripts of the US presidential debates available in the presidential project (www.presidency.ucsb.edu), the oral production of candidates can be analysed. Such a choice allows us to detect the real style and thinking mode of a policymaker. A primary medium for communication, the role and impact of the oral mode are increasing with the widespread use of the Internet with its audio and video sources.

This study focusses on the following research questions. First, can we detect some drifts in the US oral political language over a time period exceeding 60 years? Do those variations correspond to some specific nominees, or are they recurrent across almost all candidates? If past studies based on written speeches indicate a language simplification over time, can we observe a similar pattern with the oral communication mode? In particular, can we classify Trump's language as a clear outlier in political language, or it is just an extension of existing trends? Drawing from many debate transcripts, is it possible to discover stylistic features explaining the success or failure of nominees?

To achieve those objectives, this analysis will concentrate on the vocabulary complexity, language sophistication, sentence length, and various measurements grounded on selected word occurrence frequencies. In order to persuade citizens, the speaker could choose between a rational argument, or one with a higher emotional intensity. In the former case, a complex formulation would be characterised by long sentences, very rich vocabulary, and logical expression. Such a rhetorical choice usually leads to a clear and logical conclusion. On the other hand, the policymaker might opt for simpler expressions, and include both positive and negative emotions. In this case, the pessimistic effect tends to have a larger impact on the audience, and encourages people to vote (McDonald & Lenz 2008).

One must recall that the oral communication mode depicts structural features which are distinct from written or web-based forms. Speech is time-bound, transient, spontaneous, and dynamic. From a linguistic point of view, the frequency of personal pronouns is higher, while the mean sentence length is shorter. Looser constructions and filler words (e.g. um, uh, …) could appear. Additionally, several terms and expressions tend to frequently recur with occasional rephrasing, features which reduce the vocabulary richness (Crystal 2003; Yule 2020). Moreover, some verbal or lexical complex forms do not usually occur in the oral form. In this case, shorter words and simpler syntax are favoured. In a dialogue, the speaker could be interrupted without being able to finish their sentence or reasoning. Of course, the transcripts used in this study do not include the intonation, volume or tempo of the speaker's voice, nor the gestures or facial expressions.

The rest of this paper is organised as follows. Section 2 presents an overview of related topics, while Section 3 describes the corpus employed in our experiments. Section 4 exposes various methods to quantify the language complexity. Section 5 analyses the stylistic and rhetoric choice of the nominees in the presidential elections. A conclusion reports the main findings of this study.

## 2 Related Work

Different studies have analysed and compared the stylistic features of numerous chiefs of executive or congressmen (Rule et al. 2015; Gennaro & Ash, 2021). To achieve this, a selection of the most significant messages is required (e.g. *State of the Union* addresses (Savoy 2015), such as end-of-year speeches (Pauli & Tuzzi 2009; Kubát et al. 2020; Labbé et al. 2025). However, those discourses are first written before being delivered to the public. Therefore, they do not closely reflect the oral communication mode.

Based on such messages[1], various stylometric studies (Savoy 2020; Kreuz 2023) propose describing the style of each leader with a few vocabulary richness indicators. From this perspective, various studies consider the Type Token Ratio (TTR) or the ratio between the number of word-types (or the vocabulary size) and the number of tokens (Hart, 1984; Baayen, 2008; Popescu et al., 2009; Hart, 2020). Other measures such as the hapax density (percentage of words occurring once) have been proposed (Baayen 2008), or the percentage of big words (namely, words having six letters or more) (Lakoff & Wehling 2012; Hart, 2020).

In addition, a stylistic study may focus on syntactic aspects of messages such as mean sentence length (measured by the number of tokens) (Savoy, 2017). The occurrence of long sentences denotes a substantiated reasoning or detailed explanation. Even if a long sentence is required, its length is usually not conducive to easy understanding (Labbé & Savoy 2020).

Related analyses have been performed on web-mediated communications (Savoy 2018). This communication medium allows the politician to avoid going through traditional media, judged to be too critical of his/her previous actions. Moreover, this choice allows the politician to elude press conferences which would require the speaker to answer questions and comments raised by journalists. Web-based posts present additional features such as emoticons, emojis, and some slang language. One of the main differences is the use of short formulations required to catch the public's attention in the hopes that these expressions will then reappear in the media's headlines. In addition, this solution allows the integration of hyperlinks that provide images or videos to support an assertion or reinforce a strong claim.

Based on different communication forms, rhetorical aspects can be evaluated with lists of related words. For example, in the category *Symbolism*[2], Hart (1984) and Hart et al. (2013) include a list of words related to country (e.g. 'nation,' 'America'), ideology (e.g. 'democracy,' 'freedom'), or generally, political concepts and institutions (e.g. 'law,' 'government'). Other word classes include the *Blame* group, which gathers together terms such as 'angry,' 'deceptive,' and 'wrong,' while the *Communication* category includes terms such as 'declare,' 'explain,' and 'propose.' Examining US presidential speeches, Hart (1984) employs the Diction system to describe the rhetorical and stylistic differences between US presidents, from Truman to Reagan. Subsequently, Hart et al. (2013) expose the political communication variations from G. Bush to Obama, while later work identifies specific features related to Trump's presidency (Hart, 2020; 2023).

Based on a similar approach, the LIWC system (Linguistic Inquiry & Word Count) (Tausczik & Pennebaker 2010) groups words under syntactic, emotional or psychological categories. Such classes may correspond to specific grammatical categories (e.g. first person singular denoted *Self* with ‘I,’ ‘me,’ ‘mine,’ ‘my’) as well as broader classes (e.g. personal pronouns). Based on semantics, the LIWC system defines the *Cognition* process (e.g. ‘knows,’ ‘believe,’ ‘think’) or terms related to *Human* (e.g. ‘family,’ ‘woman,’ ‘child*’[3]). Using the LIWC system, Slatcher et al. (2007) were able to determine the personalities of different political candidates in the 2004 US presidential election. They defined their psychological portraits using several measurements (e.g. the relative frequency of different pronouns, emotions, etc.) and a set of composite indices reflecting cognitive complexity, personal status or honesty.

---

[1] In this study, the term 'message' implies the written form.

[2] In this study, the denomination of a wordlist is capitalized and depicted in italics.

It is important, however, to note that automatically extracting political tone or psychological traits from texts is not exempt from concerns (Grimmer & Stewart 2013). All natural languages present some lexical ambiguity with polysemic terms (e.g. ‘head’, ‘run’, ‘foot’). The short context may change the prior meaning of a term (e.g. ‘nuclear’ in ‘nuclear atomic model’), and the meaning of idioms (e.g. ‘beat around the bush’) should not be taken at face value. Moreover, not all terms or expressions in a given category correspond to the same intensity. A candidate could talk about an ‘opponent’ or ‘rival’, but using the term ‘enemy’[4] implies a stronger negative view.

Finally, a few studies have been conducted to automatically extract the semantics of a set of speeches, such as the *State of the Union* addresses (SOTU) from 1790 to 2014 (Rule et al. 2015). Based on probabilistic topic models (Blei 2012), Rule et al. (2015) show that up to 1917, the main topics of SOTUs were statecraft and political economy. After this date, the focus changed to mainly domestic and foreign policy. A similar study has been done on the speeches of Congressmen (Benoit et al. 2020). Based on a manual analysis of the SOTU and inaugural speeches, Lim (2002) found five main trends characterising more recent discourses, namely a certain anti-intellectualism, the presence of more poetic terms, a clear assertive tone, a people-oriented rhetoric, and the use of colloquial language.

## 3 Corpus

Unlike previous studies, the current corpus focuses on the oral communication mode of nominees in the presidential elections. Downloaded from the website www.presidency.ucsb.edu, this collection covers 34 debates performed by 19 nominees from 1960 to 2024. As shown in Table 1, not all exchanges of views are available, and a complete coverage starts in 1988. We must, however, remember that no debate took place during the period from 1964 to 1972. In this table, the year and number of debates are depicted in the second and third columns.

For each presidential election, one can count one to three debates that were broadcasted on TV channels and cable news. Each discussion is supervised by a moderator who introduces the main topics and poses questions to the candidates. As a variant, the moderator might let the floor address queries directly to the nominees. A rigorous comparison between nominees is therefore possible, since each is having to speak under the same conditions (e.g. similar questions, equal speaking time, same setting).

[3] When generating an entry in a wordlist, one can use the symbol ‘*’ to denote any sequence of letters.

[4] According to Levitsky & Ziblatt (2018), this change is more than an increase in intensity, and denotes an authoritarian attitude (Linz 1978). This norm of mutual tolerance in the US was also broken just before and after the Civil War (1861–1865).

**Table 1.** Statistics about our corpus with the number of debates, the transcript length and median number of sentences per speech turn

| | Year | Debate | Length | Turn Length |
|---|---|---|---|---|
| Nixon | 1960 | 3 | 14,368 | **370** |
| Kennedy | 1960 | 3 | 13,232 | 351 |
| Ford | 1976 | 3 | 13,990 | 273 |
| Carter | 1976 | 3 | 17,071 | 341 |
| Carter | 1980 | 1 | *5,681* | 204 |
| Reagan | 1980 | 2 | 10,498 | 277 |
| Dukakis | 1988 | 2 | 14,490 | 184 |
| H. Bush | 1988 | 2 | 11,760 | 193 |
| H. Bush | 1992 | 3 | 15,219 | 150 |
| Clinton | 1992 | 3 | **30,553** | 211 |
| Clinton | 1996 | 2 | 15,558 | 187 |
| Dole | 1996 | 2 | 16,511 | 190 |
| Gore | 2000 | 3 | 20,543 | 180 |
| G. Bush | 2000 | 3 | 22,186 | 138 |
| G. Bush | 2004 | 3 | 19,820 | 200 |
| Kerry | 2004 | 3 | 22,845 | 270 |
| McCain | 2008 | 3 | 20,483 | 97 |
| Obama | 2008 | 3 | 22,702 | 113 |
| Obama | 2012 | 3 | 22,327 | 63 |
| Romney | 2012 | 3 | 23,867 | 56 |
| H. Clinton | 2016 | 3 | 20,083 | 59 |
| Trump | 2016 | 3 | 23,346 | 36 |
| Trump | 2020 | 2 | 14,711 | *16* |
| Trump | 2024 | 2 | 16,863 | 154 |
| Biden | 2020 | 2 | 14,161 | 19.5 |
| Biden | 2024 | 1 | 6,774 | 171 |
| Harris | 2024 | 1 | 6,053 | 204 |

The text length per candidate is depicted in the fourth column of Table 1 where the highest value is shown in bold and the smallest in italics. In total, this corpus contains 455,695 tokens[5], corresponding to approximatively twice the length of Melville's *Moby Dick* (around 250,000 tokens). The average length over all candidates is 16,878 tokens. Clinton appears as having the largest number of tokens (1992: 30,553), followed by Romney (23,867), and Trump (2016: 23,346). When counting per candidate, Trump (having participated in three presidential campaigns and seven debates) depicts the highest value (54,920 tokens).

[5] Punctuation symbols and dollar signs are not counted as tokens.

The last column of Table 1 depicts the median number of words per speech turn. For example, the highest score is achieved by Nixon (370), indicating that he spoke 370 terms, in median, during each of his speech turns. When comparing the mean answer length per nominee we have ignored short replies composed of six words or less, which usually serve as acknowledgements in a dialogue (e.g. "Thank you." or "Yes, Joe, it's correct."). When comparing the average from Nixon to Reagan with the nominees from Obama to Trump, one can observe a clear reduction in length. For the first ones, the mean is 303, while for the second group the average is 76.5. In 2024, one can observe a rise of this score (mean: 176.7).

To obtain a better view of this evolution, Figure 1 clearly shows the underlying drift. To explain this tendency, one can signal a social demand for shorter answers, pressure that might be increased by a societal requirement, by the requirement to cover more topics in the same time frame, by the challenger or moderator[6].

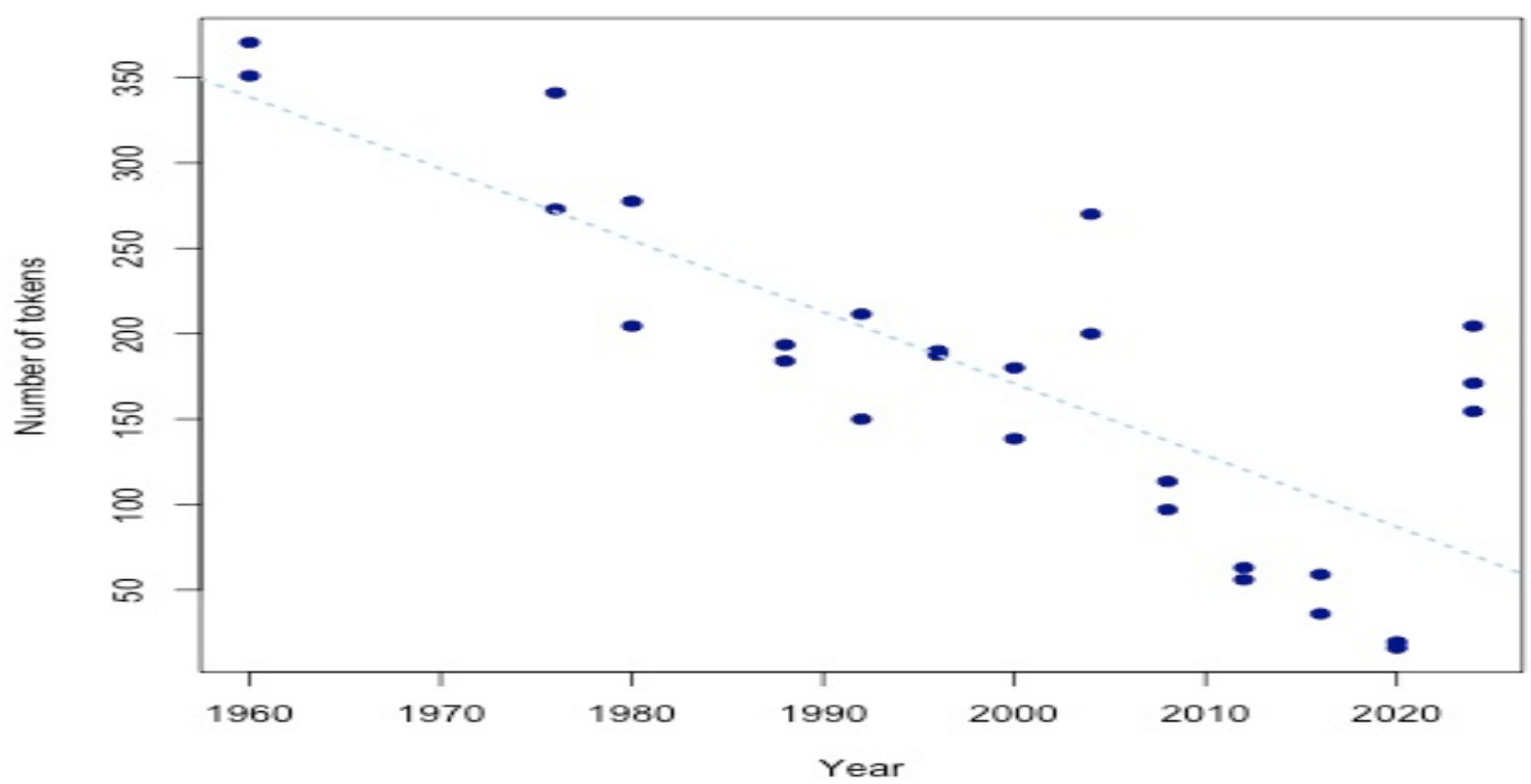

**Figure 1.** Evolution of the median length of each speech turn (number of words)

Finally, as the transcripts are extracted from a debate, one can observe that a nominee could be commenting on a claim previously expressed by the participant opponent, or indicate a fundamental opposition. For example, Dukakis was the first to interrupt his challenger (H. Bush) with an observation specifying that the previous claim was incorrect ("That's not true. That's not true.", Sept. 25th 1988). Such a spontaneous reaction reappears with Obama ("That's not true, John", Sept 26th, 2008). Later, with Trump, such interruptions are simplified and occur more often (e.g. with a simple "Wrong. Wrong", Sept, 26th, 2016). Such disturbances could explain, partially, the reduction of the speech turn.

[6] One must recall that Jim Lehrer, journalist (PBS) and novelist, was moderator for 12 debates from 1988 to 2012. Thus, the moderator cannot be the main factor explaining this decrease.

## 4 Vocabulary Complexity and Sentence Length

The most frequent method to measure lexical complexity is to rely on the TTR. This value is computed by dividing the vocabulary size (denoted by V) by the text length (indicated by n), resulting in TTR = V / n. A high TTR score signals a rich and diverse vocabulary, usually generated by an educated writer or speaker. Of course, a low value is associated with poor vocabulary diversity and repetitive lexical choices. This solution could also reflect a person who needs to insist on a reduced number of announcements or claims.

This method, however, has a main drawback. When working with texts of different lengths from the same source, these measures are not really constant and thus do not match with a distinctive or stable stylistic feature (Baayen 2001; 2008).

To avoid this problem, a better computation was suggested by Covington & McFall (2010) and Popescu et al. (2009). Instead of computing a single measure for a text, these authors suggest taking the moving average of TTR (denoted MATTR) (Kubát & Milička 2013). Based on written speeches, the MATTR varies from one US president to the next. For example, the highest score was achieved by Washington (41.2%), and the lowest by Wilson (36.1%) (Savoy 2017).

Table 2 reports the MATTR per nominee and year based on non-overlapping bins of 500 tokens. The highest score is achieved by Clinton (1992: 0.423) and the smallest by Trump (2020: 0.338). It is worth noting that, up until Clinton, the nominees display higher MATTR scores (mean: 0.404) while the last three show a smaller value (mean: 0.362). This is a first indicator that even in oral mode, the politicians tend to simplify their speech, a finding also found when analysing written speeches (Benoit et al. 2019).

A bilateral *t*-test has been applied to compare MATTR values to the average (0.387). Any statistically significant differences are shown by a † (significance level $\alpha = 5\%$). In addition, the Benjamini-Hochberg procedure (James et al. 2021) has been applied to control the false discovery rate. The same statistical inference procedure is used for the following comparisons with the mean.

As displayed in Table 2, some older and more recent candidates have MATTR values depicting a significant difference from the average. Interestingly, Harris shows a value within range of the mean, while both Trump and Biden have significantly lower MATTR scores.

As a second method to measure the lexical complexity, one can evaluate the percentage of big words (BW), defined as words composed of six letters or more (at least for the English language) (Hart 2020). For example, one can observe that some terms are easier to understand than others, such as ‘cars’ versus ‘automobiles’, or ‘TV’ versus ‘television’ (Lakoff & Wehling 2012). When a speech presents a high percentage of big words, its vocabulary is judged to be sophisticated and harder to follow and understand by the audience. For example, Fillmore (1850–1852) is the US president with the highest BW percentage (37.3%) in messages compared to Clinton (25.1%) (Savoy 2017).

**Table 2.** Language complexity computed according to the moving average TTR, the percentage of big words (BW), and mean sentence length

| | MATTR | BW (in %) | MSL |
|---|---|---|---|
| Nixon (1960) | 0.376 † | 22.5% † | 21.8 † |
| Kennedy (1960) | 0.397 | 24.1% † | 21.5 † |
| Ford (1976) | 0.398 | 25.2% † | 22.2 † |
| Carter (1976) | 0.417 † | 24.4% † | 21.0 † |
| Carter (1980) | 0.417 † | **28.1**% † | **23.9** † |
| Reagan (1980) | 0.412 † | 21.6% | 23.3 † |
| Dukakis (1988) | 0.395 | 21.4% | 20.0 † |
| H. Bush (1988) | 0.396 | 19.8% | 15.3 |
| H. Bush (1992) | 0.393 | 18.6% † | 14.2 † |
| Clinton (1992) | **0.423** † | 21.6% | 18.6 † |
| Clinton (1996) | 0.420 † | 21.2% | 19.1 † |
| Dole (1996) | 0.398 | 18.8% † | 13.9 † |
| Gore (2000) | 0.411 † | 20.8% | 17.3 |
| G. Bush (2000) | 0.387 | 20.0% | 14.3 † |
| G. Bush (2004) | 0.402 | 20.4% | 14.1 † |
| Kerry (2004) | 0.399 | 19.0% † | 15.1 † |
| McCain (2008) | 0.393 | 20.7% | 15.2 |
| Obama (2008) | 0.409 † | 20.5% | 18.6 † |
| Obama (2012) | 0.405 † | 20.6% | 18.9 † |
| Romney (2012) | 0.383 † | 19.8% | 14.9 † |
| H. Clinton (2016) | 0.408 † | 20.3% | 15.9 |
| Trump (2016) | 0.351 † | 16.8% † | 11.3 † |
| Trump (2020) | *0.338* † | *15.3*% † | *10.0* † |
| Trump (2024) | 0.354 † | 16.8% † | 10.8 † |
| Biden (2020) | 0.364 † | 16.3% † | 11.6 † |
| Biden (2024) | 0.363 † | 17.1% † | 14.2 † |
| Harris (2024) | 0.400 | 22.5% † | 16.5 |
| Mean | 0.387 | 19.2% | 16.8 |

As depicted in the third column of Table 2, the percentage of BW decreases over the years, demonstrating a tendency towards simplification of political speech. The highest score is achieved by Carter (1980: 28.1%), and the lowest by Trump (2020: 15.3%). Compared to the written mode, the percentages of BW in oral communication are clearly lower. Oral mode favours shorter words, usually to achieve a better understanding.

To compare these BW values to the mean (19.2%), a bilateral *t*-test has been computed and significant differences are displayed by a † ($\alpha = 5\%$). As shown, the two extremities of the BW values depicted scores that are always significantly different from the mean, but in the opposite direction. In this case too, the last nominees opt for a simplification of wording, favouring terms that are short and easy to understand. This method also confirms a reduction in the language complexity. For example, the mean percentage of BW from Nixon to Reagan rises to 23.6%, while this average for Trump and Biden is only 16.5%. Policymakers are clearly taking account of Lakoff & Wehling's (2012) finding that longer words are harder for an audience to understand.

As a third approach to evaluate the complexity of a speech, the number of tokens (composed of letters and/or digits) per sentence (mean sentence length, MSL) has been calculated. As shown in the last column of Table 1, these values vary clearly from one candidate to the next, with Carter depicting the highest mean (1980: 23.9 tokens/sentence). As its counterpart, Trump shows the lowest one (2020: 10).

When considering the last three candidates, the mean length of their sentences is rather low and composed, in mean, by 12 tokens. For older candidates from Nixon to Reagan, the mean sentence length is 22.3, reflecting a more elaborate explanation or reasoning. This reduction of the sentence length seems to begin with H. Bush (1988), without being a straightforward linear decrease (e.g. Clinton or Obama display a mean value higher than that of their opponents). Moreover, the *t*-test ($\alpha = 5\%$) comparing these values to the mean (16.8) confirms the significant difference between the recent candidates and the older ones. Here again, the significant differences are evolving into the opposite direction, with more recent nominees opting for a simpler syntax.

Overall, the lowest values in Table 3 occur with Trump in 2020. The three measures (MATTR, BW, MSL) indicate that Trump opted for a simplification of the political language. In 2020, however, Biden also depicts low values for these features, thus following a similar pattern. It is interesting to note that Harris (2024) had chosen a more traditional communication strategy, showing higher values for these three measures. Can we consider the higher values of those linguistic features as evidence for explaining Harris' failure?

The two runners-up present distinct styles, with Harris uttering more complex words in longer sentences. Harris' MATTR is 13% higher than Trump's value, and she appears to have similar MATTR and BW values to scores achieved by older presidents. When focussing on MSL, the democratic nominee portrays a value that is 53% higher than Trump's score. Such differences in oral communication form could partially explain Harris' failure or difficulty to be clearly understood.

## 5 Emotional and Rational Language

Our evaluation of the emotional and rational aspects has been grounded on LIWC. This text-based evaluation system is built around several wordlists according to stylistic, syntactical or psychological categories. The central assumption is to assume that words reflect the way the speaker thinks or feels (Jordan 2022). LIWC categories can be applied to measure both positive emotions (defined as *Posemo*) (e.g. 'happy,' 'hope,' 'peace') or negative ones (*Negemo*) (e.g. 'fear,' 'blame*'). The speaker's affect can be judged using these categories. Table A.1 in the Appendix depicts these two LIWC classes for each candidate and year. Populist leaders more often employ emotional terms to incite strong sentiments in the population and to obtain larger media coverage (Obradović *et al*. 2020; Hart 2020; Savoy & Wehling 2022).

Instead of focusing on a single percentage related to a given wordlist, the LIWC system proposes a combination of several categories to generate four composite measurements, namely emotional *Tone*, *Confidence* (or clout), *Analytical* thinking, and *Authenticity*. These four broad categories are suitable to analyse political communication. To convince the audience, a candidate may choose to present arguments using more emotions or affect (measured by the *Tone* category). Such a choice is usually made by Congressmen when debating patriotism, foreign policy, or social issues (Gennaro & Ash 2021). As an alternative, the speaker may decide to expose explanations in a more logical or rational shape (measured by the *Analytical* group). In addition, the politician must usually demonstrate a personality that has higher status (*Clout*), and should be viewed as an honest person (*Authenticity*).

These four stylistic measurements are standardised scores based on some LIWC categories, and their values range from 1 to 100 (Pennebaker *et al*. 2014; Jordan *et al*. 2019). The computed values obtained with our corpus are depicted in Table 3, in which the largest values appear in bold and the smallest in italics. A bilateral *t*-test has been applied and statistically significant differences are denoted by a † (significance level $\alpha = 5\%$).

The emotional *Tone* (Monzani *et al*. 2021) merges both positive and negative dimensions. Values larger than 50 indicate an overall positive tone, while numbers below this threshold are associated with a generally negative sentiment. As shown in Table 3, the positive ones dominate presidential debates, in part because the candidates must convince the citizens that they have the capacity to solve current problems, and that the country would have a bright future under their direction.

From this perspective, G. Bush presents the highest score (67.30), and Dole the second highest (63.01). On the opposite end, Trump displays the lowest value for tone (32.06), and a clearly negative one. The second lowest value is also achieved by a recent candidate (Biden, 39.94). One might believe that recent years tend to present more complex problems for political nominees, and indicate a less optimistic vision for the future.

The *Clout* (or confidence) category is employed to define the person's relative status in a social hierarchy. A leader must have a high status, usually reflected by a higher usage of the pronouns 'we' and 'you' (see also Table A.1). On the opposite end, a person of lower status tends to employ more I-words and impersonal pronouns (e.g. it, one) (Kacewitz *et al*. 2014; Pennebaker 2011). People of high social status present higher authoritative language and have a tone of higher certainty.

As depicted in Table 3, the lowest confidence value is achieved by H. Bush (63.05), who tends to opt more frequently for I-words. Low scores also characterise candidates from Nixon to Reagan. In the Table A.1 (see Appendix), H. Bush presents the highest percentage of I-words (4.69%), compared to 2.07% for Biden.

On the contrary, the last four candidates present a high score in this dimension, indicating their willingness to include the audience in their speeches (e.g. high frequency of we-words). Moreover, their wording tends to demonstrate their certainty of having the most pertinent solutions for the nation. In addition, they also need to reassure the public of their capability and their confidence in a brilliant future for the country.

**Table 3.** Composite summary measurements (LIWC) per nominee

| | *Tone* | *Clout* | *Analytical* | *Authenticity* |
|---|---|---|---|---|
| Nixon | 43.94 † | 69.72 † | 59.18 | 32.02 |
| Kennedy | 47.36 † | 67.14 † | 74.57 † | 32.30 |
| Ford | 53.82 | 69.47 † | **84.20** † | 29.77 |
| Carter | 46.50 † | 68.33 † | 79.31 † | 33.85 |
| Reagan | 47.10 † | 68.22 † | 62.77 † | 35.63 |
| Dukakis | 61.28 † | 76.78 | 57.33 | 30.74 |
| H. Bush | 56.34 | *63.05* † | 51.62 | 42.29 † |
| Clinton | 57.44 † | 74.67 | 63.96 † | 38.47 † |
| Dole | 63.01 † | 76.39 | 44.02 † | 39.27 † |
| Gore | 58.51 † | 66.97 † | 65.77 † | **46.18** † |
| G. Bush | **67.30** † | 76.37 | 60.28 | 28.94 |
| Kerry | 41.13 † | 74.09 | 58.55 | 35.79 |
| McCain | 54.32 | 77.84 | 54.13 | 26.26 † |
| Obama | 59.30 † | 80.73 † | 44.53 † | 31.06 |
| Romney | 55.55 | 74.12 | 59.14 | 37.80 † |
| H. Clinton | 57.09 | 81.35 † | 44.39 † | 21.63 † |
| Trump | *32.06* † | 82.28 † | *26.07* † | 29.38 |
| Biden | 39.94 † | **83.75** † | 46.55 † | *17.17* † |
| Harris | 56.80 | 83.58 † | 59.54 | 22.02 † |
| Mean | 52.26 | 75.73 | 54.67 | 32.22 |

The *Analytical* thinking measure has been shown to be associated with a greater academic level (Markowitz, 2023). This tone is grounded on larger cognitive elaboration, leading to an impression of conveying more competence. Analytical language appears to be logical and formal, employs more articles and prepositions, and focuses more on noun phrases (Pennebaker *et al.* 2014; Jordan *et al*. 2019). Opting for a highly analytical tone, the speaker takes the risk of appearing too distant, impersonal, and lacking in emotional timbre. On the other hand, a more intuitive and personable person speaks more often with pronouns, negations, auxiliary verbs, conjunctions and some adverbs (e.g. so, very) (Pennebaker *et al*. 2014).

In the presidential debates, Ford (84.20) presents the highest analytical thinking, followed by Carter (79.31) and Kennedy (74.57). This logical reasoning and cognitive thinking were clearly an essential attribute for candidates in 60’ and 70’. More recent politicians present lower values in this psychological trait such as, for example, the lowest value achieved by Trump (26.07), which has an important gap with the second lowest (Dole: 44.02). Current citizens tend to react more to negative emotions than logical reasoning (Gennaro & Ash 2021). When considering this factor, Trump clearly appears to be distinct from the others, and reflects this mode of thinking. Moreover, this aspect is promoted by the widespread adoption of social media by Trump’s electoral organisation as their first communication channel (Hart 2020).

The *Authenticity* measurement (Pennebaker *et al*., 2014) is related to the way a leader is able to communicate in a spontaneous way (Markowitz *et al.*, 2023), a tone usually viewed to be an honest one. Adopting this characteristic, the language is more concrete and presents more self-references in a natural way. Leaders adopting this tone appear to be closer or more connected to people's interests (Hart 2023). However, this attitude does not imply that the speaker tells the truth (Pennebaker 2011).

As displayed in Table 3, Gore (46.18) presents the highest value, with H. Bush (42.29) in second place and Dole (39.27) in third. With the lowest values, one can find Biden (17.17), H. Clinton (21.63), and Harris (22.02). In this case, Trump obviously appears to be closer to the people than his opponents are (Hart 2020).

Overall, when comparing the set of nominees from H. Clinton to Biden to the first ones (Nixon to Reagan), one can observe some clear shifts. Recent candidates are less optimistic about the nation and the future, and show some certainty and confidence in their wordings. They employ less rational thinking and explanations, and appear to be more authentic and closer to the people.

In Appendix (Table A.2), one can find the same psychological measurements subdivided by years or electoral campaigns. When considering the emotional *Tone* dimension, both H. Bush and G. Bush have a clear decrease from their first campaign to the second (1988: 62.21 vs. 51.68 in 1992; 2000: 78.71 vs. 52.45 in 2004). With Trump, the tone is rather negative during his three electoral campaigns and the lowest value was achieved in 2024. It is interesting to note that Obama displays the opposite direction (2008: 55.01 vs. 63.55 in 2012), being more positive during his second campaign.

# 6 Conclusion

Based on the oral communication of nominees during the US presidential elections, this study shows that language sophistication significantly decreases over decades. Previous researches based on written messages have shown a similar tendency (Lim 2002), (Benoit *et al*. 2019). As depicted in Table 2, the candidates present a higher rate of lexical repetition measured by MATTR scores. From this point of view, Biden and Trump show a significant reduction across the overall mean. The lexical complexity measured by the percentage of Big Words (BW) follows the same pattern. Contemporary candidates lean towards the use of short words that are easier for the audience to understand.

At the syntax level, recent nominees tend to speak with shorter sentences (in mean, 11 tokens per sentence) than older candidates (22.3 tokens) (see Table 2). This linguistic simplification signals that while substantiated reasoning or a detailed explanation were the norm in the past, they are no longer required or even desired by the audience. Additionally, shorter sentences are conducive to an easier understanding.

When evaluating the emotional and psychological traits (see Table 3), the nominees (in mean) tend to adopt a positive tone, and G. Bush depicts the highest score in this aspect. The last two presidents have, however, chosen a more negative viewpoint. During an election, the emotional aspect could play a key role in motivating voters (Brader 2005; McDonald & Lenz 2008). In this case, negative emotions such as anger and fear had been present in political advertisements for a while. The effectiveness of such a communication strategy depends on three main factors: tone, party of the supporting candidate, and topics (Ansolabehere & Iyengar 1995).

On the other hand, the candidates' confidence (Jordan *et al*. 2019) increases over the years. More so than for the older ones, current nominees present with a high degree of certainty the conviction of promoting the best political offer for the nation. In addition, they are sure of having the required capacity to assume power and provide a bright future for the country.

When evaluating the analytical thinking of the nominees, one can observe a clear change over the years. From Nixon to Carter, the candidates present a similar picture, adopting logical and rational expression to present their arguments. Among the last ones, namely H. Clinton, Trump, and Biden, the choice was to embrace a more emotional language with less logical reasoning in order to persuade citizens.

Finally, this study shows that Trump's language must not be analysed as a significant outlier in the US political world. It is more of a follow-up in clear tendencies towards language uncomplicatedness. Trump's wordings are clearly repetitive, avoid complex terms, and opt for short sentences. Those features also occur with other contemporary candidates such as Biden. Trump's formulation clearly presents a negative emotional tone, with the certainty of always knowing how to solve all problems. Moreover, Trump's language comes across as being authentic and corresponding to a person that is closely aligned with people's main interests. This linguistic feature is distinct from other nominees. Finally, Trump avoids formal and rational thinking, favouring expressions that are able to catch the audience's interest, and which might emerge as headlines in the media.

## Appendix

**Table A.1.** Results of some LIWC categories over each nominee and year

| | Posemo | Negemo | I | We |
|---|---|---|---|---|
| Nixon (1960) | 2.50% † | 1.50% † | 2.75% † | 2.98% † |
| Kennedy (1960) | 2.74% † | 1.57% | 2.80% † | 3.19% † |
| Ford (1976) | 2.79% † | 1.29% † | 2.68% † | 2.53% † |
| Carter (1976) | 2.76% † | 1.75% | 2.58% † | 2.98% † |
| Carter (1980) | 3.47% † | 2.00% † | 2.28% † | 2.36% † |
| Reagan (1980) | 2.49% † | 1.33% † | 2.86% † | 2.44% † |
| Dukakis (1988) | 3.48% † | 1.61% | 3.02% | 3.28% † |
| H. Bush (1988) | 3.25% † | 1.33% † | **4.69**% † | 2.22% † |
| H. Bush (1992) | 3.15% | 1.77% | 4.41% † | 2.36% † |
| Clinton (1992) | 3.19% | 1.58% | 3.77% † | 2.66% |
| Clinton (1996) | 3.28% † | 1.49% † | 3.67% † | 2.84% |
| Dole (1996) | 3.39% † | 1.43% † | 3.75% † | 2.75% |
| Gore (2000) | 3.32% † | 1.59% | 3.91% † | 2.22% † |
| G. Bush (2000) | **3.93**% † | *1.03*% † | 4.01% † | 2.53% † |
| G. Bush (2004) | 3.85% † | 2.42% † | 3.41% | 3.25% † |
| Kerry (2004) | 3.06% | 2.21% † | 3.51% † | 2.58% † |
| McCain (2008) | 3.46% † | 1.94% † | 3.15% | 3.05% † |
| Obama (2008) | 3.19% | 1.64% | 2.35% † | 3.73% † |
| Obama (2012) | 3.38% † | 1.39% † | 2.09% † | **3.89**% † |
| Romney (2012) | 3.16% | 1.58% | 3.03% | 2.88% |
| H. Clinton (2016) | 3.31% † | 1.66% | 3.31% | 3.15% † |
| Trump (2016) | 2.77% † | 2.28% † | 3.52% † | 2.81% |
| Trump (2020) | 2.41% † | 1.61% | 3.46% † | 2.26% † |
| Trump (2024) | 2.27% † | **2.46**% † | 2.95% | 2.73% |
| Biden (2020) | 2.48% † | 1.31% † | 2.21% † | *2.07*% † |
| Biden (2024) | *2.10*% † | 2.10% † | *2.07*% † | 2.76% |
| Harris (2024) | 3.12% | 1.48% † | 2.56% † | 2.46% † |
| Mean | 3.05% | 1.68% | 3.14% | 2.78% |

**Table A.2.** Composite summary measurements (LIWC) per candidate and year

| | *Tone* | *Clout* | *Analytical* | *Authenticity* |
|---|---|---|---|---|
| Nixon (1960) | 43.94 † | 69.72 † | 59.18 | 32.02 |
| Kennedy (1960) | 47.36 | 67.14 † | 74.57 † | 32.30 |
| Ford (1976) | 53.82 | 69.47 † | 84.20 † | 29.77 |
| Carter (1976) | 44.25 † | 69.25 † | 76.75 † | 33.65 |
| Carter (1980) | 53.42 | 65.53 † | **86.00** † | 34.42 |
| Reagan (1980) | 47.10 | 68.22 † | 62.77 † | 35.63 † |
| Dukakis (1988) | 61.28 † | 76.78 | 57.33 | 30.74 |
| H. Bush (1988) | 62.21 † | *61.00* † | 53.47 | 43.31 † |
| H. Bush (1992) | 51.68 | 64.65 † | 50.19 | 41.50 † |
| Clinton (1992) | 56.31 | 73.60 | 64.07 † | 39.76 † |
| Clinton (1996) | 59.63 † | 76.66 | 63.75 † | 35.96 † |
| Dole (1996) | 63.01 † | 76.39 | 44.02 † | 39.27 † |
| Gore (2000) | 58.51 † | 66.97 † | 65.77 † | **46.18** † |
| G. Bush (2000) | **78.71** † | 71.88 † | 61.28 | 31.62 |
| G. Bush (2004) | 52.45 | 80.84 † | 59.13 | 26.11 † |
| Kerry (2004) | 41.13 † | 74.09 | 58.55 | 35.79 † |
| McCain (2008) | 54.32 | 77.84 | 54.13 | 26.26 |
| Obama (2008) | 55.01 | 78.45 † | 48.48 † | 31.75 |
| Obama (2012) | 63.55 † | 82.91 † | 40.56 † | 30.34 |
| Romney (2012) | 55.55 | 74.12 | 59.14 | 37.80 † |
| H. Clinton (2016) | 57.09 † | 81.35 † | 44.39 † | 21.63 † |
| Trump (2016) | 34.32 † | 81.91 † | *25.53* † | 32.57 |
| Trump (2020) | 40.19 † | 84.24 † | 25.83 † | 30.14 |
| Trump (2024) | *22.84* † | 80.99 † | 27.07 † | 24.61 † |
| Biden (2020) | 47.42 | 82.04 † | 44.76 † | 17.20 † |
| Biden (2024) | 25.77 † | **86.96** † | 50.37 | *17.08* † |
| Harris (2024) | 56.80 | 83.58 † | 59.54 | 22.02 † |
| Mean | 51.40 | 75.06 | 55.59 | 31.83 |

## References

Ansolabehere, S., & Iyengar, S. (1995). *Going negative. How political advertisements shrink & polarize the electorate.* The Free Press: New York.

Baayen, H.R. (2001). *Word Frequency Distribution*. Kluwer : Dordrecht

Baayen, H.R. (2008). *Analyzing Linguistic Data. A Practical Introduction Using R*. Cambridge: Cambridge University Press.

Benoit, K., Munger, K., & Spirling, A. (2019). Measuring and Explaining Political Sophistication Through Textual Complexity. *American Journal of Political Science*, 63(2):491–508.

Blei, D. M. (2012). Probabilistic Topic Models. *Communication of the ACM*, 55(4):77-84

Brader, T. (2005). Striking a Responsive Chord: How Political Ads Motivate and Persuade Voters by Appealing to Emotions. *American Journal of Political Science,* 49(2):388–405.

Covington, M.A., & McFall, J.D. 2010. Cutting the Goridan Knot: The Moving-Average Type-Token Ratio (MATTR). *Journal of Quantitative Linguistics*, 17(2):94–100.

Crystal, D. (2003). *The Cambridge Encyclopedia of the English Language*. Cambridge: Cambridge University Press.

Gennaro, G., & Ash, E. (2021). Emotion and Reason in Political Language. *The Economic Journal,* 132, 1037–1059.

Grimmer, J., & Stewart, B. M. (2013). Text as Data: The Promise and Pitfalls of Automatic Content Analysis Methods for Political Texts. *Political Analysis*, 21(3):267–297.

Hart, R. P. (1984). *Verbal Style and the Presidency. A Computer-based Analysis*. New York: Academic Press.

Hart, R.P., Childers, J.P., & Lind, C.J. (2013). *Political Tone. How Leaders Talk and Why*. Chicago: The University of Chicago Press.

Hart, R.P. (2020). *Trump and Us. What He Says and Why People Listen*. Cambridge: Cambridge University Press.

Hart, R.P. (2023). *American Eloquence: Language and Leadership in the Twentieth Century*. New York: Columbia University Press.

James, G., Witten, D., Hastie, T. & Tibshirani, R. (2021). *An Introduction to Statistical Learning with Applications in R.* New York: Springer.

Jordan, K.N., Sterling, J., Pennebaker, J.W., & Boyd, R.L. (2019). Examining Long-Term Trends in Politics and Culture Through Language of Political Leaders and Cultural Institutions. *Proc. National Academy of Science,* 116(9):3476–3481.

Jordan, K. (2022). Language Analysis in Political Psychology. In *Handbook of Language Analysis in Psychology,* M. Dehghani, R.L. Boyd (eds), 159-172. New York: Guilford Publications.

Kacewicz, E., Pennebaker, J. W., Davis, M., Jeon, M., & Graesser, A. C. (2014). Pronoun use reflects standings in social hierarchies. *Journal of Language and Social Psychology,* 33(2):125–143.

Kreuz, R. (2023). *Linguistics Fingerprints. How Language Creates and Reveals Identity*. Prometheus Books.

Kubát, M., & Milička, J. (2013). Vocabulary Richness Measures in Genres. *Journal of Quantitative Linguistics*, 20(4):339–349.

Kubát, M., Macutek, J., & Cech, R. (2020). Communists Spoke Differently: An Analysis of Czechoslovak and Czech Annual Presidential Speeches. *Digital Scholarship in the Humanities*, 36(1),:138-152.

Labbé, D., Labbé, C., & Savoy, J. (2025). ChatGPT as Speechwriter for the French Presidents. *Digital Scholarship in the Humanities*. Advance online publication. https://doi.org/10.1093/llc/fqaf032

Lakoff, G., & Wehling, E. (2012). *The Little Blue Book: The Essential Guide to Thinking and Talking Democratic*. New York, Free Press.

Levitsky, S., & Ziblatt, D. (2018). *How Democracies Die. What History Reveals About our Future*. New York: Crown Press.

Lim, E.T. (2002). Five Trends in Presidential Rhetoric: An Analysis of Rhetoric from George Washington to Bill Clinton. *Presidential Studies Quarterly*, 32(2) :328–366.

Linz, J. J. (1978). *The Breakdown of Democratic Regimes*. Baltimore, John Hopkins University Press.

Markowitz, D.M. (2023). Analytic Thinking as Revealed by Function Words: What does Language Really Measure? *Applied Cognitive Psychology,* 37(3):1–8.

Markowitz, D.M., Kouchaki, M., Gino, F., Hancock, J.T., & Boyd, R.L. (2023). Authentic First Impressions Relate to Interpersonal, Social, and Entrepreneurial Success. *Social Psychological and Personality Science,* 14(2):107–116.

McDonald Ladd J., & Lenz, G.S. (2008). Reassessing the Role of Anxiety in Vote Choice. *Political Psychology*, 29(2):275–296.

Monzani, D., Vergani, L., Pizzoli, S. F. M., Marton, G., & Pravettoni, G. (2021). Emotional Tone, Analytical Thinking, and Somatosensory Processes of a Sample of Italian Tweets during the First Phases of the COVID-19 Pandemic: Observational study. *Journal of Medical Internet Research, 23*(10), e29820–e29820.

Obradović, S., Power, S.A., & Sheehy-Skeffington, J. (2020). Understanding the Psychological Appeal of Populism. *Current Opinion in Psychology,* 35(10):125–131.

Pauli, F., & Tuzzi, A. (2009). The End of Year Addresses of the Presidents of the Italian Republic (1948–2006): Discourse Similarities and Differences. *Glottometrics*, 18, 40–51.

Pennebaker, J.W. (2011). *The Secret Life of Pronouns. What our Words Say About Us*. New York: Bloomsbury Press.

Pennebaker, J.W., Chung, C.K., Frazee, J., Lavergne, G.M., & Beaver, D.I. (2014). When Small Words Foretell Academic Success: The Case of College Admissions Essays. *PLoS ONE* 9, doi.org/10.1371/journal.pone.0115844

Popescu, I.-I., Atlman, G., Grzybek, P., Jayaram, B. D., Köhler, R., Krupa, V., Macutek, J., Pustet, R., Uhlirova, L., & Vidya, M.N. (2009). *Word Frequency Studies*. Berlin: De Gruyter Mouton.

Rule, A., Cointet, J. P., & Bearman, P. S. (2015). Lexical Shifts, Substantive Changes, and Continuity in State of the Union Discourse, 1790-2014. *Proceedings National Academy of Sciences* (PNAS), 112(35).

Savoy, J. (2017). Analysis of the Style and the Rhetoric of the American Presidents over Two Centuries. *Glottometrics*, 38:55–76.

Savoy, J. (2018). Analysis of the Style and the Rhetoric of the 2016 US Presidential Primaries. *Digital Scholarship in the Humanities*, 33(1):143–159.

Savoy, J. (2020). *Machine Learning Methods for Stylometry. Authorship Attribution and Author Profiling*. Cham: Springer.

Savoy, J., & Wehren, M. (2022). Trump's and Biden's Styles During the 2020 US Presidential Election. *Digital Scholarship in the Humanities*, 37(1):229–241.

Savoy, J. (2025). Stylometric analysis of inaugural, *State of the Union*, and farewell addresses. *International Journal of Digital Humanities*, 7:511–536.

Slatcher, R. B., Chung, C. K., Pennebaker, J. W., & Stone, L. D. (2007). Winning Words: Individual Differences in Linguistic Style Among U.S. Presidential and Vice-Presidential Candidates. *Journal of Research in Personality*, 41(1): 63–75.

Tausczik, Y.R., & Pennebaker, J.W. (2010). The Psychological Meaning of Words: LIWC and Computerized Text Analysis Methods. *Journal of Language and Social Psychology*, 29(1):24–54.

Yule, G. (2020). *The Study of Language*. Cambridge: Cambridge University Press